\documentclass[lettersize,journal]{IEEEtran}
\usepackage{amsmath,amsfonts}
\usepackage{algorithmic}
\usepackage{algorithm}
\usepackage{array}
\usepackage[caption=false,font=normalsize,labelfont=sf,textfont=sf]{subfig}
\usepackage{textcomp}
\usepackage{stfloats}
\usepackage{url}
\usepackage{verbatim}
\usepackage{graphicx}
\usepackage{cite}
\usepackage{booktabs}
\usepackage{caption}
\usepackage{longtable}
\usepackage{threeparttable}
\usepackage[numbers,sort&compress]{natbib}
\usepackage{hyperref}
\usepackage{booktabs}
\usepackage{array}
\usepackage{color}

\begin{document}

\title{Appformer: A Novel Framework for Mobile App Usage Prediction Leveraging Progressive Multi-Modal Data Fusion and Feature Extraction}

\author{Chuike Sun, Junzhou Chen, Yue Zhao, Hao Han, Ruihai Jing, Guang Tan, and Di Wu
\thanks{Corresponding author: Junzhou Chen.}
\thanks{Chuike Sun, Junzhou Chen, Yue Zhao, and Guang Tan are with School of Intelligent Systems Engineering, Shenzhen Campus of Sun Yat-sen University, No. 66, Gongchang Road, Guangming District, Shenzhen, Guangdong 518107, P.R. China (e-mail: sunchk3@mail2.sysu.edu.cn; chenjunzhou@mail.sysu.edu.cn; zhaoy376@mail2.sysu.edu.cn; tanguang@mail.sysu.edu.cn).}
\thanks{Hao Han and Ruihai Jing are with Transsion Holdings Co., Ltd, Shanghai 201203, China(e-mail: hao.han@transsion.com; ruihai.jing@transsion.com).}
\thanks{Di Wu is with the School of Computer Science and Engineering, Sun Yat-sen University, Guangzhou, 510006, China, and also with the Guangdong Key Laboratory of Big Data Analysis and Processing, Guangdong 510006, China (e-mail: wudi27@mail.sysu.edu.cn). }
}

\maketitle

\begin{abstract}
This article presents Appformer, a novel mobile application prediction framework inspired by the efficiency of Transformer-like architectures in processing sequential data through self-attention mechanisms. Combining a Multi-Modal Data Progressive Fusion Module with a sophisticated Feature Extraction Module, Appformer leverages the synergies of multi-modal data fusion and data mining techniques while maintaining user privacy. The framework employs Points of Interest (POIs) associated with base stations, optimizing them through comprehensive comparative experiments to identify the most effective clustering method. These refined inputs are seamlessly integrated into the initial phases of cross-modal data fusion, where temporal units are encoded via word embeddings and subsequently merged in later stages. The Feature Extraction Module, employing Transformer-like architectures specialized for time series analysis, adeptly distils comprehensive features. It meticulously fine-tunes the outputs from the fusion module, facilitating the extraction of high-calibre, multi-modal features, thus guaranteeing a robust and efficient extraction process. Extensive experimental validation confirms Appformer's effectiveness, attaining state-of-the-art (SOTA) metrics in mobile app usage prediction, thereby signifying a notable progression in this field.

\end{abstract}

\begin{IEEEkeywords}
App Usage Prediction, Transformer, Multi-Modal Data Fusion, Feature Extraction, Data Mining.
\end{IEEEkeywords}

\section{Introduction}
\IEEEPARstart{I}{n} today's digital era, mobile applications have become integral to people's lives, enabling interactions in various daily activities\cite{akdim2022role,zhang2020app,zhu2013mobile,deng2020optimal,li2020extent,xie2021trimming,li2021understanding,huang2017shuffledog}. By analyzing interactive information, we can learn the behaviors and preferences of different users, aid in understanding user needs, and provide personalized services to improve user experience. Accurately predicting users' app usage has become a prominent area of research\cite{lu2022machine,natarajan2013app,baeza2015predicting,shin2012understanding,tu2019personalized}, forming the foundation for personalized recommendation systems and significantly impacting mobile app development and user experience enhancement.

However, predicting mobile app usage behavior remains a formidable challenge, as it is influenced by a multitude of factors including time, spatial context, and individual preferences. The associated data often display attributes like brevity, high dimensionality, discreteness, and multi-modality. Traditional prediction methodologies frequently struggle with issues such as data sparsity, abstract feature extraction, and maintaining accuracy in model predictions. Recently, deep learning techniques, particularly those employing attention mechanisms, have emerged as a significant advancement in the realm of mobile application prediction, offering effective solutions to these intricate challenges.

In the realm of mobile app prediction, our research endeavors to address three pivotal challenges:
\begin{enumerate}
    \item \textbf{Accurately representing core data:} This involves user IDs, spatiotemporal information (time and location of app usage), and app usage history sequences. We meticulously encode these components to capture their nuances.
    \item \textbf{Efficiently integrating multimodal data:} We employ advanced multimodal data fusion techniques to seamlessly integrate the encoded data, fostering a synergistic synthesis of information.
    \item \textbf{Enhancing feature extraction:} We aim to empower prediction models to distill insightful and discriminative features from the unified dataset, crucial for accurate and robust predictions.
\end{enumerate}

To address these challenges, our research presents innovative solutions aimed at enhancing the prediction efficiency for mobile application usage. The key contributions are summarized as follows:
    
\begin{itemize}
    \item \textbf{Appformer Framework:} We introduce the Appformer framework, leveraging attention mechanisms and Transformer architectures to significantly enhance data fusion and feature extraction. It comprises two synergistic components: a Multi-Modal Data Progressive Fusion Module and a Feature Extraction Module. The former adeptly combines encoded data from diverse sources using cross-modal fusion technology, setting a strong base for data synthesis. The latter, with its Encoder-Decoder architecture, efficiently extracts temporal features from time-series data. Adjustments to the Encoder and Decoder inputs have been made to optimize multi-modal feature extraction. Together, these modules boost the framework's ability to deliver accurate and efficient application predictions.

    \item \textbf{Multi-modal Data Progressive Fusion:} At the heart of Appformer's fusion module is a Cross-Modal Data Fusion Module, built on cross-modal and self-attention techniques. It can seamlessly integrate data from application sequences, Points of Interest (POIs), user IDs, and time, each treated with a tailored, phased fusion process. This approach ensures data integrity, fosters data synthesis, and significantly improves feature extraction and analysis quality.

    \item \textbf{Enhanced Preprocessing of Location and Time Data:} Acknowledging the critical role of location and time data in mobile application prediction, we have advanced their preprocessing. For location data, we prioritize user privacy, selecting base station-related POI data, and apply clustering to refine its representation. This optimized data feeds into the early stages of our fusion process. Time data undergoes word embedding and concatenation before joining the fusion later on, ensuring a comprehensive data integration.

    \item \textbf{Validation through Extensive Experiments:} Our experiments affirm Appformer's state-of-the-art performance on several metrics. Tested across different data processing and partitioning approaches on a unified dataset, our framework achieved Hit@1 scores of 31.92\% and 42.68\%. These scores represent improvements of 4.75\% and 7.89\% over existing methods, showcasing our proposed framework's superior capability in predicting mobile application usage.
\end{itemize}

\section{Related Work}
The application prediction problem involves using the historical information of application usage to forecast the potential applications that users may use in the future. Through literature research, we have established a strong correlation between time information, user information, location information, and application usage. Based on this, we have decomposed the application prediction problem into three parts: mining and expression of core data, fusion of multi-modal data, and feature extraction network. Each of these parts has been elaborated on in detail.

{\bf{Correlation analysis between core data and application prediction:}} Wang \textit{et al.}\cite{wang2019modeling} found not only a strong correlation between the locations accessed by users on their mobile devices and the applications they use but also discovered that users tend to use different apps during various time periods. Nadai \textit{et al.}\cite{de2019strategies} significant disparities were observed in how different users engage with apps. Moreover, user app preferences were found to evolve over time. Xia \textit{et al.}\cite{xia2020deepapp} analyzed a large-scale real-world dataset of app usage and found that smartphone app usage exhibits spatio-temporal correlation and personalization. Garrido \textit{et al.}\cite{graells2018and} discovered that the physical locations within a user's city can influence their app usage patterns significantly. Tian \textit{et al.}\cite{tian2021and} found that users with different characteristics have different app preferences and that users use different apps during various time periods. These studies demonstrate a strong correlation between spatio-temporal information, user information, and application usage. We need to make rational use of these data for app prediction.

{\bf{Mining and expression of core data:}} App prediction involves data encoding, similar to word embedding in Natural Language Processing (NLP). Word embedding is a significant research area, converting words into high-dimensional vectors to enhance the understanding of natural language. One-hot encoding is a simple and easy-to-implement method, but it fails to capture semantic relationships\cite{selva2021review}. The Word2Vec model proposed by Mikolov et al.\cite{mikolov2013efficient} was a major breakthrough, introducing efficient training algorithms. The GloVe model proposed by Pennington et al.\cite{pennington2014glove} further improved word embeddings by combining global and local information. FastText, proposed by Joulin et al.\cite{joulin2016bag}, utilizes deep learning methods and can handle both word-level and subword-level information. We can construct data processing methods for relevant data in app prediction based on the word embedding methods in NLP.

{\bf{Fusion of multi-modal data:}} In the field of application prediction, the integration of encoded multi-modal data has become essential. Multi-modal data fusion is a critical process that helps capture the correspondence between different types of contextual information data, such as user personalized features, space, time, and historical app sequences, reducing information redundancy and enhancing generalization. It can significantly improve the accuracy and robustness of application prediction models.

In recent years, there has been a surge in research and applications related to multi-modal data fusion. Performing multi-modal data fusion requires building models that are adept at handling and associating information from multiple sources\cite{li2018review}\cite{baltruvsaitis2018multimodal}. Traditional methods for multi-modal data fusion include co-training algorithms\cite{brefeld2004co,muslea2000selective,yang2012information}, co-regularization algorithms\cite{kan2015multi}\cite{sun2011multi}, marginal consistency algorithms\cite{sun2013multi}\cite{chao2016consensus}, and multiple kernel learning (MKL)\cite{gonen2011multiple}.

Furthermore, more complex techniques have been used for multi-modal data fusion. These include variants of deep Boltzmann machines for modeling the joint distribution of different modalities\cite{srivastava2012multimodal}\cite{hu2013multimodal}, extensions of classical autoencoders for discovering correlations between hidden representations of two modalities\cite{wang2015deep}\cite{feng2014cross}, nonlinear extensions of canonical correlation analysis (CCA) using deep neural networks\cite{andrew2013deep}, and architectures based on convolutional neural networks (CNN) that excel at combining information from multiple sources\cite{su2015multi,yao2017deepsense,wang20172d}.

In addition, groundbreaking approaches have emerged, further enriching the field of application prediction. For example, the Multi-modal Transformer (MulT) technique introduced by Tsai \textit{et al.}\cite{tsai2019multimodal}, the Multi-modal Temporal Graph Attention Network (MTGAT) proposed by Yang \textit{et al.}\cite{yang2020mtag}, spatial fusion encoding developed by Wang \textit{et al.}\cite{wang2022am3net}, the Multi-modal Attention based Feature Fusion (MAT) by Islam \textit{et al.}\cite{islam2020hamlet}, and the Modality-Invariant Cross-modal Attention (MICA) proposed by Liang \textit{et al.}\cite{liang2021attention}.

\begin{figure}[t]
\captionsetup{justification=centering}
\centering
\includegraphics[width=0.5\textwidth]{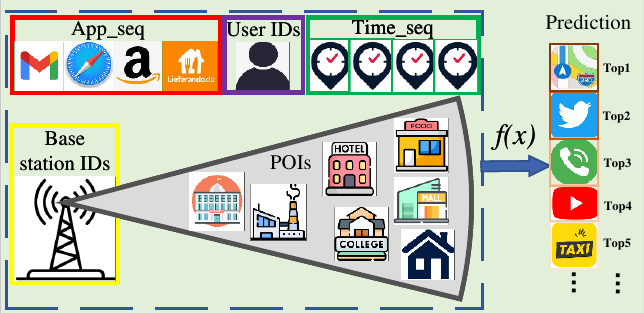}
\caption{Problem definition and relevant feature data for app usage prediction.}
\label{fig_1}
\end{figure}

{\bf{Feature extraction network:}} Feature extraction network is a type of network structure widely used in deep learning techniques in recent years. Here are some examples of related feature extraction networks:

Lee \textit{et al.}\cite{lee2019app} employed a stacked Long-Short Term Memory (LSTM) architecture, a sequence-based deep learning framework utilized for training prediction models without the need to calculate transition probabilities. Zhao \textit{et al.}\cite{zhao2019appusage2vec} introduced a novel framework for predicting application usage called AppUsage2Vec, inspired by Doc2Vec. This framework models application usage records by taking into account the contributions of various applications, personalized user features, and temporal context. By incorporating an application attention mechanism, it evaluates the impact of each application on the target application. User-specific features in application usage are learned using a dual-DNN (Deep Neural Network) module. Xia \textit{et al.}\cite{xia2020deepapp} presented a new framework for predicting application usage known as DeepApp. This framework accomplishes context-aware prediction through the use of multi-task learning\cite{ruder2017overview}. Ouyang \textit{et al.}\cite{ouyang2022learning} proposed a new model, the dynamic usage graph network (DUGN), to explicitly model complex app correlations using a dynamic graph structure for learning user interest dynamics. Moreira \textit{et al.}\cite{moreira2020nap} developed a prediction model called Natural App Processing (NAP), which is inspired by language modeling (LM). The NAP deep learning framework is composed of LSTM layer, dropout layer, and softmax layer, which are used to make application prediction. Furthermore, Solomon \textit{et al.}\cite{solomon2022predicting} proposed a deep learning framework for predicting application usage. This framework includes Gate Recurrent Unit (GRU), attention layer, and softmax layer. Shen \textit{et al.}\cite{shen2019deepapp} introduced a deep reinforcement learning framework, DeepAPP, that learns a model-free predictive neural network from past app usage data. They also devised an online updating strategy for adapting the network to fluctuating app usage behavior.

{\bf{Limitations of existing work:}} Through comprehensive literature research, we have identified shortcomings in current prediction methods. Firstly, regarding the representation of core data, existing methods often lack the necessary depth and breadth in processing such data, failing to fully and accurately reveal the true implications of the data. This could lead to misinterpretation and misuse of data, thereby affecting the accuracy of predictions. Secondly, concerning data fusion, current strategies for data fusion are often too simplified or overly complex, failing to properly handle and integrate multimodal data. This could result in the loss of important information or interference from irrelevant information, reducing prediction accuracy. Lastly, from the perspective of feature extraction networks, existing networks often have limited capabilities, unable to effectively extract useful features from raw data. This could lead to inaccurate and unstable predictions, impacting user experience. These issues have severely limited the performance and effectiveness of current methods in the field of application prediction. Therefore, we need to build a new system to address these problems and improve the accuracy of predictions and user experience.

\begin{figure*}[!t]
\captionsetup{justification=centering}
\centering
\includegraphics[width=\linewidth]{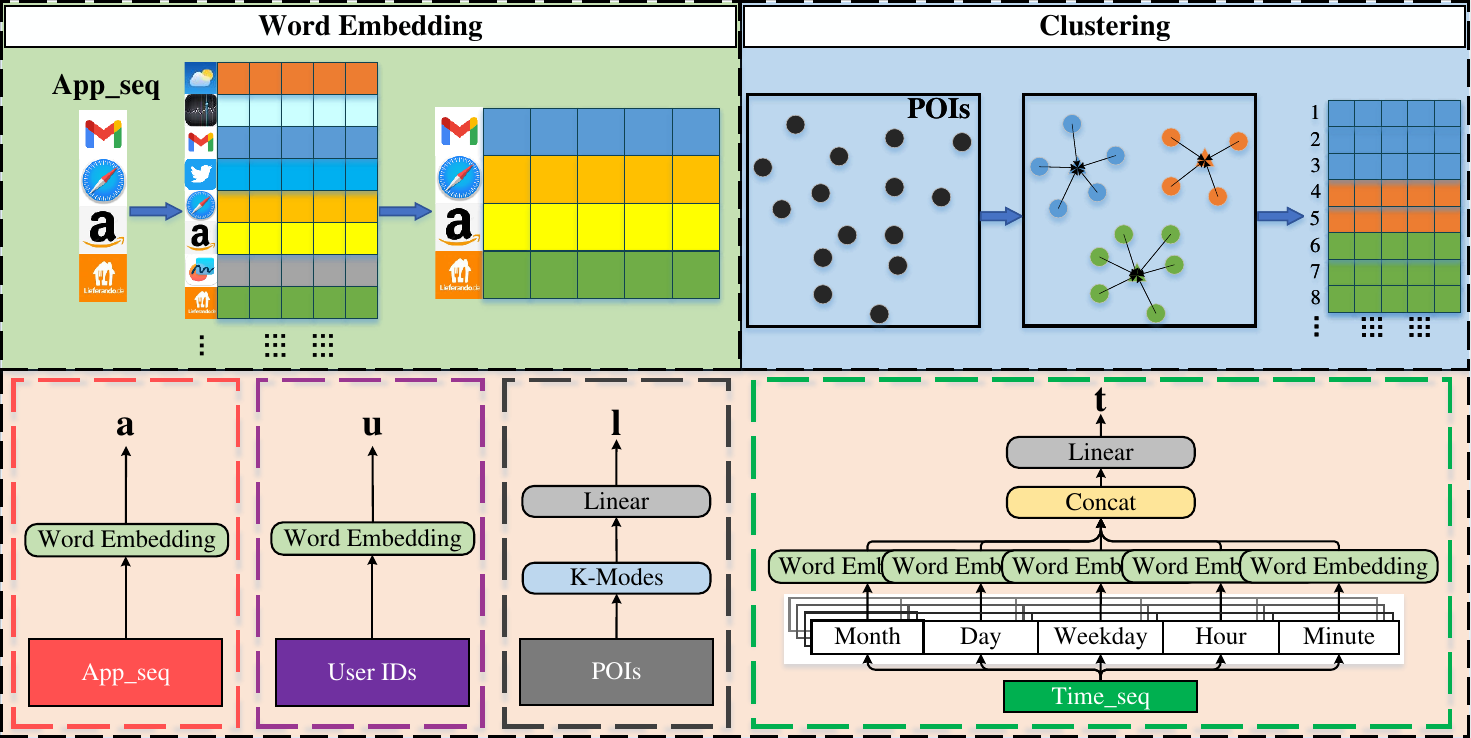}
\caption{Data preprocessing for app usage prediction.}
\label{fig_2}
\end{figure*}

\section{Problem Description And Dataset}
\subsection{Problem Description}
As illustrated in Figure \ref{fig_1}, our dataset consists of various feature data types. The application sequence is highlighted in a red box (App\_seq), user IDs in a purple box (User IDs), the timestamp sequence of application starts in a green box (Time\_seq), the base station ID interacting with the mobile phone during application usage in a yellow box (Base station IDs), and the Points of Interest (POIs) data corresponding to the base station's coverage sector in a grey sector box (POIs). The POI data, indicating the presence of different facilities near a base station, plays a vital role in defining the functionality and appeal of specific locations \cite{yu2018smartphone}.

Our goal is to develop a predictive model $f(x)$ capable of precisely predicting the next application a user is likely to use. This model is designed to rank the likelihood of potential next apps in a descending order, starting with the most probable choice as Top1, followed by the next most likely options as Top2, Top3, and so forth. The mathematical representation is illustrated by the following equation:
\begin{equation}
\label{eq_1}
A(n+1) = f\left(\mathbf{X}(n-m+1:n)\right)
\end{equation}

where $\mathbf{X}(n-m+1:n)$ represents a multi-dimensional feature sequence, incorporating essential historical data required for prediction. This encompasses sequences of app IDs, user IDs, timestamps, base station IDs, and POI vectors, seamlessly integrated into a coherent structure for the predictive model. The configuration of this sequence is defined as follows:
\begin{equation}
\label{eq_2}
\begin{aligned}
&\mathbf{X}(n-m+1:n) = \\
&\left\{ (A(i), U(i), T(i), B(i), P(i)) \mid i = n-m+1, \ldots, n \right\}
\end{aligned}
\end{equation}
This structured approach allows our model to effectively leverage the rich, interconnected data, enhancing its predictive accuracy for upcoming application usage.

\subsection{Dataset}
\label{Dataset}
Our study employs a real-world application usage dataset collected from Shanghai, China, during a week-long period from April 20 to April 26, 2016 \cite{yu2018smartphone}. Each dataset entry contains an anonymous user ID, a timestamp, a base station ID, and the ID of the app being used, organized chronologically by timestamp. Table \ref{tab_1} provides a comprehensive view of the dataset format, featuring five sequential records from a single user for illustrative purposes.

Moreover, every base station in the dataset is designated by a unique ID and corresponds to a 17-dimensional Points of Interest (POI) vector. These POI vectors reflect the socioeconomic activities characteristic of the areas around the base stations, encompassing categories such as Medical care, Hotels, and Business affairs. In our analysis, POI data plays a crucial role in enhancing the prediction of app usage patterns. Table \ref{tab_2} displays the POIs linked to the three base stations mentioned in Table \ref{tab_1}, providing insight into the local functionalities and services available in their vicinity.

\begin{table}[h]
\centering
\caption{Examples of dataset.}
\label{tab_1}
\begin{tabular}{cccc}
\toprule
User ID & Timestamp & Base Station ID & App ID \\
\midrule
723 & 20160420151912 & 7327 & 2 \\
723 & 20160420151915 & 9419 & 237 \\
723 & 20160420151921 & 9617 & 1 \\
723 & 20160420151941 & 7327 & 2 \\
723 & 20160420151942 & 7327 & 2 \\
\bottomrule
\end{tabular}
\end{table}

\begin{table*}[!t]
  \centering
    \caption{Examples of POIs.} \label{tab_2}
        \begin{tabular}{*{9}{c}}
        \toprule
        Base Station ID & Medical care & Hotel & Business affairs & Life service & Transportation hub\tnote{1} & Culture & Sports & Residence \\ \midrule
        7327 & 0 & 0 & 9 & 1 & 0 & 0 & 0 & 2  \\ 
        9419 & 0 & 0 & 25 & 0 & 0 & 0 & 0 & 0 \\ 
        9617 & 0 & 2 & 0 & 1 & 1 & 0 & 0 & 35 \\ 
        \midrule
        Entertainment and leisure\tnote{2} & Scenic spot & Government & Factory & Shopping & Restaurant & Education & Landmark & Other \\ \midrule
        2 & 1 & 0 & 0 & 11 & 13 & 2 & 0 & 0 \\ 
        0 & 0 & 1 & 4 & 0 & 0 & 0 & 2 & 0 \\ 
        2 & 5 & 0 & 4 & 5 & 0 & 0 & 1 & 0 \\ 
        \bottomrule
        \end{tabular}
\end{table*}

\section{Method}
In this section, we begin by exploring the methodologies employed for data selection and preprocessing, focusing on the implementation of word embeddings, the application of clustering techniques to Points of Interest (POIs) data, and the encoding of temporal data. Following this preliminary discussion, we introduce the Appformer framework, offering a comprehensive overview and then proceeding to elaborate on its two primary components: the Multi-Modal Data Progressive Fusion Module and the Feature Extraction Module.

\subsection{Data selection and preprocessing}
\label{Data preprocessing}
To respect and protect user privacy, we have chosen POI information associated with base stations as the input for location features. So our core data includes four parts: app sequence, user IDs, POI data and time data. The preprocessing of data are illustrated in Figure \ref{fig_2}.

\subsubsection{App sequence} As depicted in the top-left green block of Figure \ref{fig_2}, we transform the discrete app sequence (App\_seq) into a continuous vector space using word embedding techniques \cite{ghannay2016word}, a method proficient in converting discrete data into continuous vectors. As shown in the red box in the orange block of Figure \ref{fig_2}, the application sequence is converted to \textbf{a} through word embedding.

We utilize the nn.Embedding algorithm in the PyTorch framework to implement word embedding. Initially, we need to initialize an embedding matrix of size $(n, d)$. Each row of the embedding matrix corresponds to an app's embedding vector. This matrix can be randomly initialized at first, and then updated during the training process via backpropagation.

The specific embedding process can be represented with the following equation:

\begin{equation}
\label{deqn_ex1b}
\begin{aligned}
E = \text{nn.Embedding}(n, d)
\end{aligned}
\end{equation}

where $E$ denotes the embedding matrix, $n$ represents the total number of apps, and $d$ is the dimension of the embedding vector.

Subsequently, for each app in the sequence, we can find the corresponding embedding vector from the embedding matrix using its ID. This process can be represented by the following equation:

\begin{equation}
\label{deqn_ex1b}
\begin{aligned}
v_i = E(i)
\end{aligned}
\end{equation}

where $v_i$ is the embedding vector of app $i$, and $E(i)$ denotes the operation of finding the embedding vector of app $i$ from the embedding matrix $E$.

In this process, the embedding matrix $E$ is a learnable parameter and will be updated during model training. Ultimately, each app is mapped to a $d$-dimensional vector, which is its embedding vector. This vector can capture the similarities and differences between apps, as well as their contextual information within the sequence.

\subsubsection{User IDs} As shown in the purple box in the orange block of Figure \ref{fig_2}, similar to the application sequences, the User IDs are also processed into \textbf{u} using word embedding.

\subsubsection{POIs} As shown in the blue block in the top right corner of Figure \ref{fig_2}, we perform clustering operations on POIs and replace each POI vector with the vector of its corresponding cluster center. We investigated several clustering algorithms that can be efficiently applied to POI data, including K-Means, Mini Batch K-Means, K-Means++, K-Modes, and K-Harmonic Means. Below, we provide a detailed comparison of these clustering algorithms:

K-Means\cite{ahmed2020k}, a well-known clustering algorithm, aims to partition data points into a predefined number of clusters by iteratively updating cluster centers. However, K-Means is highly sensitive to the initial selection of cluster centers and can converge to local optima. To expedite convergence, we utilized the Mini Batch K-Means method\cite{hicks2021mbkmeans}, which performs iterative updates on small batch data samples, making it suitable for large datasets. This method reduces computation time while preserving accuracy. To address the issue of initial cluster center selection, we adopted the K-Means++ method\cite{li2022collaborative}, which optimizes the selection of initial cluster centers, thereby accelerating convergence to the global optimum and enhancing the stability and efficiency of the algorithm. To accommodate the discrete features of POI data, we employed the K-Modes algorithm\cite{sharma2015k}, which effectively manages a large quantity of discrete data and successfully uncovers patterns in clusters. This is particularly beneficial for processing discrete POI vectors. Lastly, considering the impact of outliers on clustering results, we introduced the K-Harmonic Means method\cite{srivastava2020computer}, which employs the harmonic mean to compute cluster centers, thus offering more robust handling of outlier data.

From the comparative experiments in Section \ref{Clustering experiments}, it can be seen that K-Modes achieved the best result. As shown in the grey box in the orange block of Figure \ref{fig_2}, we first use K-Modes to cluster the POIs, then use a linear layer to perform a linear transformation on them, obtaining the location data \textbf{l}.

The following is the specific operation of clustering POIs using K-Modes:

We denote the base station ID dataset as $X=\{x_1, x_2, ..., x_n\}$, where each $x_i=(x_{i1}, x_{i2}, ..., x_{im})$ represents a data point, i.e., a 17-dimensional POI vector of a base station ID.

Before the iteration process, we need to initialize the modes $Z=\{z_1, z_2, ..., z_k\}$. One common way to do this is to randomly select $k$ data points from the dataset $X$ as the initial modes.

Let $Z=\{z_1, z_2, ..., z_k\}$ be the current set of modes, where each $z_j=(z_{j1}, z_{j2}, ..., z_{jm})$ is a mode. The dissimilarity measure or mode distance $d(x_i, z_j)$ is defined as the number of mismatches between $x_i$ and $z_j$, i.e., 

\begin{equation}
\label{deqn_ex1a}
\begin{aligned}
d(x_i, z_j) = \sum_{q=1}^{m} \delta(x_{iq}, z_{jq})\\
\end{aligned}
\end{equation}

where $\delta(x_{iq}, z_{jq})$ is a mismatch function defined as follows:

\begin{equation}
\label{deqn_ex1b}
\begin{aligned}
\delta(x_{iq}, z_{jq}) = 
\begin{cases} 
0, & \text{if } x_{iq} = z_{jq} \\
1, & \text{otherwise}
\end{cases}
\end{aligned}
\end{equation}

In each iteration, for each $x_i$, find the mode $z_j$ that is nearest to it, i.e., find $j$ such that $d(x_i, z_j)$ is minimized, and assign $x_i$ to $z_j$.

Update each mode $z_j$ to be the mode of the data points assigned to it, i.e., for each $q=1,2,...,m$, set 

\begin{equation}
\label{deqn_ex1c}
\begin{aligned}
z_{jq} = \text{mode}(\{x_{iq} | x_i \text{ is assigned to } z_j\})
\end{aligned}
\end{equation}

where mode() denotes the mode.

This process continues until the modes no longer change or a predefined number of iterations is reached.

After the clustering process, we obtain the final set of modes (or cluster centers) $Z=\{z_1, z_2, ..., z_k\}$. Each mode $z_j=(z_{j1}, z_{j2}, ..., z_{jm})$ represents a cluster center vector.

For each data point $x_i$, we replace its original 17-dimensional interest point vector with the cluster center vector of the cluster to which $x_i$ is assigned. This is done as follows:

For each $i=1,2,...,n$, we set $x_i = z_j$, where $z_j$ is the cluster center of the cluster to which $x_i$ belongs in the final clustering result.

After this replacement, we can use these new vectors for further operations.

\subsubsection{Time data} As shown in the green box in the orange block of Figure \ref{fig_2}, we extract information such as month, day, weekday, hour, and minute from the time data (Time\_seq). Each extracted feature is then separately subjected to Word embedding, and the resulting vectors are concatenated. These concatenated vectors undergo a further linear transformation to obtain the time information matrix, denoted as \textbf{t}. We compare this approach with conventional encoding methods\cite{zhou2021informer}, where time information is encoded and combined with relative positional encoding. In our approach, we have fused the encoded time information with other relevant data in the later stage of cross-modal data fusion, rather than merely adding it to the position encoding. The results of this comparison are presented in Section \ref{Time encoding comparison experiment}.

After data preprocessing, the data transformed into:
\begin{equation}
\label{deqn_ex1a}
\begin{aligned}
\mathbf{a}=\text{WE}(\text{App\_seq})
\end{aligned}
\end{equation}

where WE denotes word embedding.

\begin{equation}
\label{deqn_ex1a}
\begin{aligned}
\mathbf{u}=\text{WE}(\text{User}\text{ }\text{IDs})
\end{aligned}
\end{equation}
\begin{equation}
\label{deqn_ex1a}
\begin{aligned}
\mathbf{l}=\text{Linear}(\text{K-Modes}(\text{POIs}))
\end{aligned}
\end{equation}
\begin{equation}
\label{deqn_ex1a}
\begin{aligned}
\mathbf{t}=&\text{Linear}(\text{Concat}(\text{WE}(\text{Month}),\text{WE}(\text{Day}),\\
&\text{WE}(\text{Weekday}),\text{WE}(\text{Hour}),\text{WE}(\text{Minute}))
\end{aligned}
\end{equation}

\subsection{Appformer Framework Overview}

\begin{figure}[t]
\captionsetup{justification=centering}
\centering
\includegraphics[width=\linewidth]{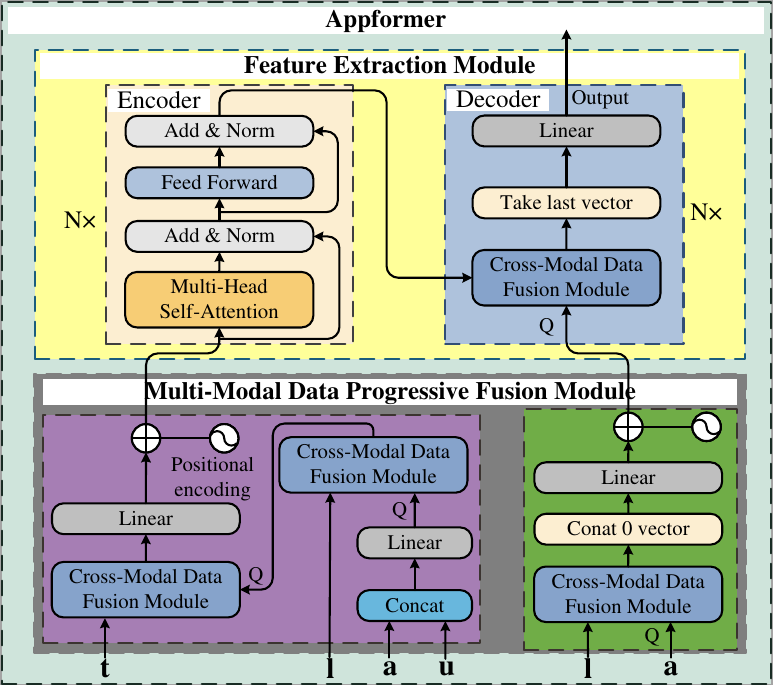}
\caption{Appformer framework for app usage prediction.}
\label{fig_3}
\end{figure}

The Appformer framework, designed with the specific aim of predicting app usage sequences, leverages multi-modal data to its advantage. As illustrated in Figure \ref{fig_3}, the Appformer comprises two core components: the Multi-Modal Data Progressive Fusion Module and the Feature Extraction Module. These components are intricately designed to work seamlessly together, first by progressively integrating data from diverse sources and then by extracting relevant features to enhance the predictive model's accuracy. To improve transparency regarding the computational process of Appformer, we have provided pseudocode that elaborates on its complete computational steps, as shown in Algorithm \ref{Algorithm_2}.

\begin{algorithm*}[!t]
\caption{Appformer: Integrated Multi-Modal Data Progressive Fusion Module and Feature Extraction Module}
\label{Algorithm_2}
\begin{algorithmic}[1]
\REQUIRE Encoded vectors $u$, $a$, $l$, $t$, model parameters
\ENSURE Final output vector $Y_{\text{pred}}$

\STATE // Multi-Modal Data Progressive Fusion Module

\STATE $fused\_app\_user \leftarrow \text{LinearTransform}(Concat(a, u))$ // Simple fusion of app and user vectors
\STATE $fused\_poi \leftarrow \text{Cross-ModalDataFusionModule}(fused\_app\_user, l)$ // Cross-modal data fusion with POI vector
\STATE $fused\_time \leftarrow \text{Cross-ModalDataFusionModule}(fused\_poi, t)$ // Cross-modal data fusion with time vector
\STATE $\textbf{encoder\_input} \leftarrow \text{AddPositionalEncoding}(\text{LinearTransform}(fused\_time))$ // Final preparation for Encoder input
\STATE $fused\_poi\_decoder \leftarrow \text{Cross-ModalDataFusionModule}(a, l)$ // Preparing Decoder input with app vector and POI vector
\STATE $decoder\_input\_prep \leftarrow \text{Concatenate}(fused\_poi\_decoder, \text{ZeroVector}())$
\STATE $\textbf{decoder\_input} \leftarrow \text{AddPositionalEncoding}(\text{LinearTransform}(decoder\_input\_prep))$

\STATE // Feature Extraction Module

\STATE $SelfAttentionOutput \leftarrow \text{MultiHeadSelfAttention}(\textbf{encoder\_input})$ // Encoder Processing
\STATE $ResidualSelfAttention \leftarrow \text{LayerNorm}(SelfAttentionOutput + encoder\_input)$
\STATE $EncoderFeedForwardOutput \leftarrow \text{FeedForward}(ResidualSelfAttention)$
\STATE $EncoderOutput \leftarrow \text{LayerNorm}(EncoderFeedForwardOutput + ResidualSelfAttention)$
\STATE $Q_{decoder} \leftarrow \textbf{decoder\_input}$ // Decoder Processing
\STATE $K_{decoder}, V_{decoder} \leftarrow EncoderOutput, EncoderOutput$ // Use EncoderOutput as K and V
\STATE $CrossModalAttentionOutput \leftarrow \text{Cross-ModalDataFusionModule}(Q_{decoder}, K_{decoder}, V_{decoder})$
\STATE $LastVector \leftarrow \text{TakeLastVector}(CrossModalAttentionOutput)$
\STATE $Y \leftarrow \text{LinearTransform}(LastVector)$

\RETURN $Y_{\text{pred}}$
\end{algorithmic}
\end{algorithm*}

\begin{table*}[!t]
\centering
% \color{red}
\caption{Detailed Comparison of Data Fusion Methods for Mobile App Prediction}
\label{tab:detailed_comparison}
% \begin{tabular}{ccc}
\begin{tabular}{>{\centering\raggedright}m{2cm}>{\centering\raggedright}m{2cm}>{\centering\raggedright\arraybackslash}m{3.5cm}>{\centering\raggedright\arraybackslash}m{9cm}}
\toprule
Method & Year & Data Fusion Method & Applicability and Limitations of Data Fusion  \\
\midrule
NAP\cite{moreira2020nap} \\ DeepApp\cite{xia2020deepapp} \\  PAULCI\cite{solomon2022predicting} & Appl. Sci 2020 \\ ACM TIST 2020 \\ Comput Commun 2022 & Heterogeneous raw data is directly concatenated and subsequently encoded, followed by additional operations.
% Various types of raw data are simply spliced together and then encoded and other operations are performed. 
& \begin{itemize}
\item Simple, efficient, and straightforward implementation.
\item Overlooks the complexity of data relationships and relevance.
\item Prone to information overload with minimal insights.
\item Scalability issues hinder performance in complex analyses.
\end{itemize} \\
\addlinespace
AppUsage2Vec\cite{zhao2019appusage2vec} & ICDE 2019 & Integrates user and app data through a Hadamard product, aiming to capture the intricate interactions between user behavior and app usage patterns.
& \begin{itemize}
\item Directly captures interactions between user behavior and app usage.
\item Element-wise multiplication may overlook complex data interactions.
\item Fails to account for dynamic user preference changes.
\item Dimensionality constraints necessitate additional preprocessing.
\item Scalability challenges impact computational efficiency.
\end{itemize} \\
\addlinespace
DUGN\cite{ouyang2022learning} & IEEE TMC 2023 & Utilizes an Interest Fusion Layer to capture the dynamic evolution of user preferences.
& \begin{itemize}
\item Adapts to real-time, evolving user interests for personalized recommendations.
\item Difficulties in managing interest changes without adding noise.
\item Requires advanced techniques to distinguish between transient and enduring interest changes.
\item Balancing evolving and current interests is challenging, risking recommendations becoming static or erratic.
\item Dependence on singular data mode may lead to incomplete user profiles.
\end{itemize} \\
\bottomrule
\end{tabular}
\end{table*}

\subsection{Multi-Modal Data Progressive Fusion Module}

In the pursuit of advancing mobile app prediction, our comprehensive review of existing data fusion methods revealed limitations in their capacity to integrate multi-modal data effectively and to discern the importance of varying data types (as shown in Table \ref{tab:detailed_comparison}). To address these shortcomings and enhance predictive performance, we propose a series of strategic improvements: emphasizing context relevance, delineating the importance of data modalities, adopting cross-modal attention mechanisms, and implementing progressive fusion strategies. These improvements have culminated in the development of the Multi-Modal Data Progressive Fusion Module, a novel approach designed to navigate the challenges identified.

As illustrated in the gray section of Figure \ref{fig_3}, the Multi-Modal Data Progressive Fusion Module, with its core Cross-Modal Data Fusion Module, cleverly merges encoded app vector \(\textbf{a}\), user vector \(\textbf{u}\), POI vector \(\textbf{l}\), and time vector \(\textbf{t}\) step by step, progressively preparing the data for the Feature Extraction Module's Encoder and Decoder.

\begin{figure*}[!t]
\centering
\captionsetup{justification=centering}
\includegraphics[width=0.75\linewidth]{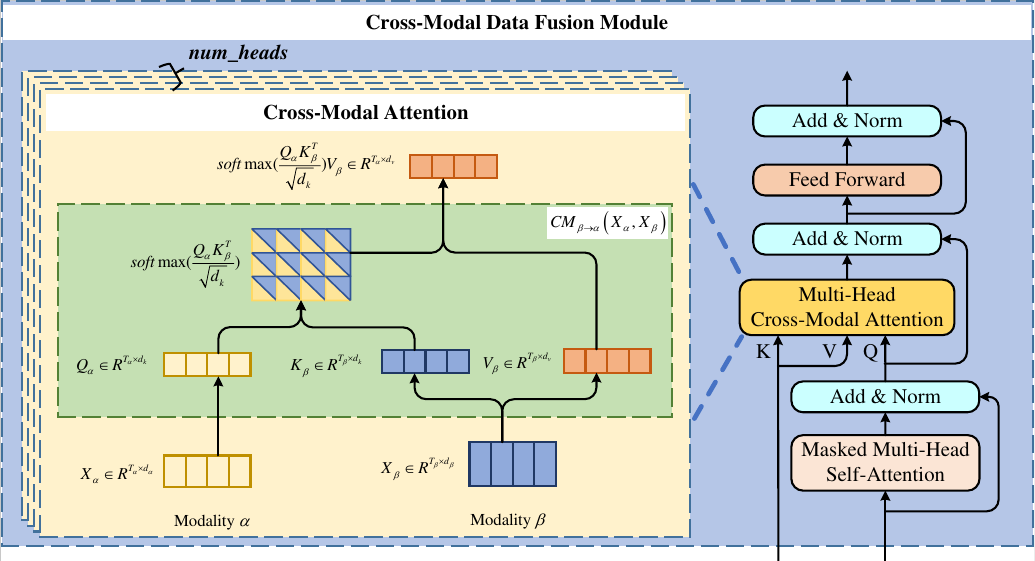}
\caption{Cross-Modal Attention and Cross-Modal Data Fusion Module for multi-modal app usage data.}
\label{fig_4}
\end{figure*}

{\bf{Cross-Modal Data Fusion Module:}} Inspired by the decoder module in the Transformer proposed by Vaswani et al.\cite{vaswani2017attention}, we integrated operations such as Multi-Head Cross-Modal Attention, Masked Multi-Head Self-Attention, Residual Connection and Layer Normalization, Feed Forward to form the Cross-Modal Data Fusion Module, as shown in Figure \ref{fig_4}. To enhance clarity around the computational workflow of Cross-Modal Data Fusion Module, we have included pseudocode detailing its computational steps, as seen in Algorithm \ref{Algorithm_1}.

\begin{algorithm*}[!t]
\caption{The Cross-Modal Data Fusion Module algorithm is exemplified by the fusion of encoded app vector $a$ and POI vector $l$.} \label{Algorithm_1}
\begin{algorithmic}[1]
\REQUIRE $a$, $l$, $d_{\text{model}}$, $num\_heads$, $dropout\_rate$, $d_{\text{ff}}$
\ENSURE $Y_{\text{fused}}$

\STATE // Initialize weight matrices for Multi-Head Self-Attention and Cross-Modal Attention
\FOR{$i = 1$ to $\text{num\_heads}$}
    \STATE $W_{Q_a}^i, W_{K_a}^i, W_{V_a}^i \leftarrow \text{InitializeWeights}(d_{\text{model}}, d_k)$ // $d_k = d_{\text{model}} / \text{num\_heads}$ % Dimension: (d_model, d_k)
    \STATE $W_{Q_l}^i, W_{K_l}^i, W_{V_l}^i \leftarrow \text{InitializeWeights}(d_{\text{model}}, d_k)$  % Dimension: (d_model, d_k)
    \STATE $W_O^i \leftarrow \text{InitializeWeights}(d_{\text{model}} \times \text{num\_heads}, d_{\text{model}})$  % Dimension: (d_model*num_heads, d_model)
\ENDFOR

\STATE // Masked Multi-Head Self-Attention for input $a$
\FOR{$i = 1$ to $\text{num\_heads}$}
    \STATE $Q_a^i \leftarrow a \cdot W_{Q_a}^i$  % Dimension of Q_a^i: (1, d_k)
    \STATE $K_a^i \leftarrow a \cdot W_{K_a}^i$  % Dimension of K_a^i: (1, d_k)
    \STATE $V_a^i \leftarrow a \cdot W_{V_a}^i$  % Dimension of V_a^i: (1, d_k)
    \STATE $MaskedAttentionWeights^i \leftarrow \text{Softmax}\left(\frac{Q_a^i \cdot (K_a^i)^\top}{\sqrt{d_k}} \cdot \text{Mask}\right)$  % Softmax over dimension (1, 1)
    \STATE $MaskedSelfAttention^i \leftarrow MaskedAttentionWeights^i \cdot V_a^i$  % Dimension: (1, d_k)
    \STATE $MaskedSelfAttention^i \leftarrow \text{Dropout}(MaskedSelfAttention^i, dropout\_rate)$  % Apply dropout
\ENDFOR
\STATE $\begin{aligned}[t]ConcatSelfAttention \leftarrow \text{Concatenate}(MaskedSelfAttention^1, \dots, MaskedSelfAttention^{\text{num\_heads}})\end{aligned}$  % Dimension: (1, d_model)
\STATE $SelfAttentionOutput_a \leftarrow \text{LayerNorm}(ConcatSelfAttention \cdot W_O + a)$  % Dimension: (1, d_model)

\STATE // Multi-head cross-modal attention using $SelfAttentionOutput_a$ as $Q$ and $l$ as $K$, $V$
\FOR{$i = 1$ to $\text{num\_heads}$}
    \STATE $Q_{\text{cross}}^i \leftarrow SelfAttentionOutput_a \cdot W_{Q_a}^i$  % Dimension: (1, d_k)
    \STATE $K_l^i \leftarrow l \cdot W_{K_l}^i$  % Dimension of K_l^i: (1, d_k)
    \STATE $V_l^i \leftarrow l \cdot W_{V_l}^i$  % Dimension of V_l^i: (1, d_k)
    \STATE $AttentionWeights^i \leftarrow \text{Softmax}\left(\frac{Q_{\text{cross}}^i \cdot (K_l^i)^\top}{\sqrt{d_k}}\right)$  % Softmax over dimension (1, 1)
    \STATE $CrossModalAttention^i \leftarrow AttentionWeights^i \cdot V_l^i$  % Dimension: (1, d_k)
    \STATE $CrossModalAttention^i \leftarrow \text{Dropout}(CrossModalAttention^i, dropout\_rate)$  % Apply dropout
\ENDFOR

\STATE $\begin{aligned}[t]ConcatCrossModalAttention 
\leftarrow \text{Concatenate}(CrossModalAttention^1, \dots, CrossModalAttention^{\text{num\_heads}})\end{aligned}$  % Dimension: (1, d_model)
\STATE $CrossModalOutput \leftarrow \text{LayerNorm}(ConcatCrossModalAttention \cdot W_O + SelfAttentionOutput_a)$  % Dimension: (1, d_model)

\STATE // Feed forward network
\STATE $W_{1} \leftarrow \text{InitializeWeights}(d_{\text{model}}, d_{\text{ff}})$  % Dimension: (d_model, d_ff)
\STATE $W_{2} \leftarrow \text{InitializeWeights}(d_{\text{ff}}, d_{\text{model}})$  % Dimension: (d_ff, d_model)
\STATE $FFNOutput \leftarrow \text{ReLU}(CrossModalOutput \cdot W_{1})$  % Apply ReLU
\STATE $FFNOutput \leftarrow \text{Dropout}(FFNOutput, dropout\_rate)$  % Apply dropout after ReLU
\STATE $FeedForwardOutput \leftarrow \text{LayerNorm}(FFNOutput \cdot W_{2} + CrossModalOutput)$  % Dimension: (1, d_model)

\STATE // Final output after another residual connection and layer normalization
\STATE $Y_{\text{fused}} \leftarrow \text{LayerNorm}(FeedForwardOutput + CrossModalOutput)$  % Dimension: (1, d_model)

\RETURN $Y_{\text{fused}}$
\end{algorithmic}
\end{algorithm*}

\subsubsection{Data Fusion for Encoder} 
As illustrated in the purple box within the gray section of Figure \ref{fig_3}, considering that we segment data by user, where each app sequence corresponds to a series of repetitive numbers for User IDs, we start by concatenating the user vector \(\textbf{u}\) with the app vector \(\textbf{a}\), followed by a simple fusion through a linear transformation. Next, the fused vector is used as the query (Q), with the POI vector \(\textbf{l}\) serving as both key (K) and value (V), to perform cross-modal fusion using the Cross-Modal Data Fusion Module, thereby integrating location information. Subsequently, the fused vector, now as the Q, with the time vector \(\textbf{t}\) as both K and V, undergoes another round of cross-modal fusion with the Cross-Modal Data Fusion Module to incorporate temporal details. Then, after performing a linear transformation and positional encoding on the fused vector, it serves as the input for the encoder of the feature extraction module.

\subsubsection{Data Fusion for Decoder} 

As illustrated in the green box within the gray section of Figure \ref{fig_3}, given that the Decoder of the Feature Extraction Module requires the most critical information to decode the output from the Encoder (which will be detailed in the introduction of the Feature Extraction Module), we therefore designate the app vector \(\textbf{a}\) as Q, and the POI vector \(\textbf{l}\) as both K and V, conducting cross-modal fusion through the Cross-Modal Data Fusion Module. We concatenate the fused vector with a zero vector, which is specifically used for storing predicted values, and then proceed with linear transformation and positional encoding, making it the input for the Decoder of the Feature Extraction Module.

\subsection{Feature Extraction Module}

\begin{table*}[!t]
\centering
% \color{red}
\caption{Detailed Comparison of Feature Extraction Methods for Mobile App Prediction}
\label{tab:detailed_comparison2}
\begin{tabular}{>{\raggedright}m{2cm}>{\raggedright}m{2cm}>{\raggedright\arraybackslash}m{4.5cm}>{\raggedright\arraybackslash}m{8cm}}
\toprule
Method & Year & Feature Extraction Method & Applicability and Limitations of Feature Extraction \\
\midrule
NAP\cite{moreira2020nap} & Appl. Sci 2020 & LSTMs.
& \begin{itemize}
\item Excel at handling sequential data.
\item Fall short in processing multimodal data, missing complex interactions.
\item The linear approach of LSTMs struggles to adapt to dynamic relevancies and changes in application importance within sequences.
\end{itemize} \\
\addlinespace
DeepApp\cite{xia2020deepapp}  & ACM TIST 2020 & Uses multiple GRUs to analyze concatenated data, aiming to understand time patterns and user behavior.
& \begin{itemize}
\item More efficient and use fewer resources than LSTMs for sequential predictions.
\item Struggle with capturing deep temporal dependencies in certain scenarios.
\item Prone to overlooking critical spatiotemporal interaction features.
\end{itemize} \\
\addlinespace
PAULCI\cite{solomon2022predicting} & Comput Commun 2022 & Combines GRU networks with an Attention Layer to selectively focus on the most informative parts of the data sequence, enhancing prediction accuracy.
& \begin{itemize}
\item Simple and straightforward, easy to deploy and optimize.
\item Applying attention mechanisms too simplistically misses key features in complex multimodal inputs.
\item The fixed structure hampers adjustments or improvements for new requirements.
\end{itemize} \\
\addlinespace
AppUsage2Vec\cite{zhao2019appusage2vec} & ICDE 2019 & Dual Deep Neural Networks (Dual-DNNs) employ two separate DNNs to independently extract user and application features before merging them.

& \begin{itemize}
\item Allows for specialized processing of user and application data.
\item May overlook complex multimodal interactions, leading to potential loss of context and inefficiency.
\item Dual-DNNs are modular but lack adaptability.
\end{itemize} \\
\addlinespace
DUGN\cite{ouyang2022learning} & IEEE TMC 2023 & Utilizes a combination of temporal gating mechanisms, hierarchical graph attention networks, and self-attention mechanisms to deeply understand user interests and their evolution.
& \begin{itemize}
\item Effectively captures user preferences.
\item Performance relies on data quality.
\item Adaptability is limited by fixed structure.
\item Ignoring multimodal data can reduce accuracy.  
\end{itemize} \\
\bottomrule
\end{tabular}
\end{table*}

After conducting an in-depth analysis of existing feature extraction methods for mobile application data (as shown in Table \ref{tab:detailed_comparison2}), we identified limitations in their ability to maintain the richness of information in complex data environments and a lack of flexibility to adapt to new data types or evolving tasks. These limitations hindered the models' ability to capture subtle relationships within the data, thereby constraining their predictive capabilities. To overcome these challenges and enhance predictive performance, we proposed a series of improvements, including adopting an Encoder-Decoder architecture to extract features that reflect the complex relationships within the data, introducing attention mechanisms to prioritize important features while reducing noise, and designing modular and replaceable feature extraction methods to improve model performance and adaptability. Leveraging these enhancements, we crafted an Feature Extraction Module as depicted in the yellow segment of Figure \ref{fig_3}, adeptly tackling the previously mentioned challenges. 

\subsubsection{Encoder and Decoder Architecture}

The Feature Extraction Module adopts the classic Encoder-Decoder architecture inspired by the Transformer architecture proposed by Vaswani et al.\cite{vaswani2017attention}. In this framework, the Encoder's primary responsibility is to receive fused multi-modal data and transform it into a complex high-dimensional representation. This transformation leverages a combination of Multi-Head Self-Attention and Feed Forward networks to intricately capture the dependencies within the data. The Decoder, on the other hand, is tasked with decoding this high-dimensional representation and combining it with an additional conditional vector to produce the target sequence. The interaction between the Encoder and Decoder is facilitated through the Cross-Modal Data Fusion Module. We take the vector from the last dimension, which is specifically used for storing predicted values, and after applying a linear transformation, use it as the final output.

\subsubsection{Differentiating Between Encoder and Decoder Inputs}

Given the distinct functions of the Encoder and Decoder, it is crucial to input the complete dataset into the Encoder to obtain a richer and more comprehensive feature vector. For the Decoder, selectively providing key information aids in more effective decoding, avoiding the interference of redundant information. According to the results from the Section \ref{Ablation experiment of Appformer decoder input} providing the Decoder with a fused vector of app and POI vectors yields the best results.

\subsubsection{Modular Design Analysis}

A key feature of our Feature Extraction Module is its modularity, which allows for easy swapping of its components with those from other well-known architectures. We conducted experiments by replacing corresponding parts of our Feature Extraction Module with modules from AutoFormer, FEDformer, and Quatformer. This was done to test the flexibility and potential improvements that could be achieved with our current architecture. The outcomes of this comparative analysis will be detailed in Section \ref{Feature extraction module comparison experiment}. This modular design not only proves the adaptability of our model but also paves the way for future enhancements by incorporating more sophisticated modules as they become available.

\section{Experiments}

\subsection{Evaluation metrics}
We use four distinct metrics to evaluate the effectiveness of different methods: Hit@k (Hit rate at k), MRR@k (Mean Reciprocal Rank at k), NDCG@k (Normalized Discounted Cumulative Gain at k) and F1 score with a macro setting.

\subsection{Baselines}

In our study, we conduct comparisons with several classic and advanced methods to validate the performance of our proposed framework. These methods include traditional machine learning classifiers as well as various deep learning models. 

In our experiments, we compare our method with the traditional machine learning classifiers: random forest\cite{breiman2001random}, logistic regression\cite{werbos1974beyond}, XGBoost\cite{chen2016xgboost}. We also compare our method with the deep learning methods:

{\bf{MLP (Multi-Layer Perceptron):}} This model utilized a multi-layer perceptron with 100 neurons, employed the $ReLU$ activation function, and used a softmax layer for application prediction.

{\bf{LSTM (Long Short-Term Memory)\cite{hochreiter1997long}:}} This model employed the LSTM architecture, capable of handling long and short sequences, and used a softmax layer for predicting the next application.

{\bf{GRU (Gate Recurrent Unit)\cite{Cho2014LearningPR}:}} Similar to LSTM, GRU could process long and short sequences and utilized a softmax layer for application predictions.

{\bf{PCA(Principal Component Analysis) and LSTM\cite{xu2018predicting}:}} This method applied PCA to spatial, application, and temporal information, which was then fed into two LSTM models. Finally, a softmax layer was used for predicting the next application.

{\bf{NAP(Natural App Processing)\cite{moreira2020nap}:}} The NAP framework consisted of two LSTM layers, a Dropout layer, and a softmax layer to provide application prediction.

{\bf{AppUsage2Vec\cite{zhao2019appusage2vec}:}} This model represented applications, time, and users as latent vectors, employed an attention network to capture the importance of each application, and used deep neural networks for application predictions.

{\bf{DeepApp\cite{xia2020deepapp}:}} DeepApp utilized multitask learning, mapping time, location, and applications to embeddings. It used a combination of GRU and linear layers for application predictions, along with tasks for user identification and location prediction.

{\bf{PAULCI\cite{solomon2022predicting}:}} This approach focuses on capturing spatial and temporal contexts using graph embeddings and multimodal embeddings. It incorporates a GRU, attention layer, and softmax layer to predict application usage, showcasing superior performance compared to traditional methods.

{\bf{DUGN\cite{ouyang2022learning}:}} DUGN leverages a dynamic graph structure to capture intricate app correlations and learn user interest dynamics. It first extracts user interests from each app usage graph using a hierarchical graph attention mechanism. The model then tracks these interests over time, creating dynamic user embeddings by modeling temporal dependencies across multiple app usage graphs. Finally, it identifies current user interests within the present app usage graph, fuses multiple interests, and generates comprehensive user embeddings for future app recommendations.

\subsection{Experimental setup}

\subsubsection{Model parameter settings}

\begin{itemize}
    \item \textbf{Sequence Length Setting:} In determining the optimal sequence length for our model, we conducted a thorough review of existing literature in the field. Based on findings from relevant studies\cite{xia2020deepapp}\cite{zhao2019appusage2vec}\cite{solomon2022predicting}, we set the parameter \(m\) in Equation \ref{eq_1} to 4, meaning that the prediction of the next app usage is based on the information related to the most recent four apps used by the user.

    \item \textbf{Embedding Dimensions:} Drawing inspiration from the work of Yin \textit{et al.} \cite{yin2018dimensionality}, we tailored the embedding dimensions for user, time, and app sequences. Specifically, we set the embedding dimension for user sequences at 16, for app sequences at 64, and for each time unit at 12, which includes month, day, weekday, hour and minute.

    \item \textbf{Configuration of Cluster Centers for POI Data:} We set the number of cluster centers in the K-Modes clustering algorithm for POI data clustering to 5. The specific determination of the number of cluster centers will be detailed in Section \ref{Clustering experiments}.
    
    \item \textbf{Framework Configuration:} In Algorithm \ref{Algorithm_1}, we configure the parameters as follows: model dimension ($d_{\text{model}}$) is set to 128, number of attention heads ($num\_heads$) to 8, dropout rate ($dropout\_rate$) to 0.05 and Feed Forward network dimension ($d_{\text{ff}}$) to 512. Moreover, we set \(N=2\) in Figure \ref{fig_3}, indicating that the Feature Extraction Module comprises 2 encoder blocks and 2 decoder blocks.

    \item \textbf{Training Regimen:} We set the training epochs to 20, batch size to 128, and used an Adam optimizer with a learning rate of 0.001 to update model parameters. Experiments are implemented on a GPU by PyTorch.
\end{itemize}

\subsubsection{Dataset cleaning and partitioning} Through literature review, we have identified the latest and optimal results on this dataset, which are presented in two papers PAULCI\cite{solomon2022predicting} and DUGN\cite{ouyang2022learning}, published in Computer Communications and IEEE Transactions on Mobile Computing, respectively. These two papers adopted different data cleaning and partitioning methods:
\begin{itemize}
    \item Solomon et al. \cite{solomon2022predicting} (PAULCI) method aggregates records where the same user stayed at the same base station and continuously used the same application for a period exceeding one minute. Sort the records in chronological order based on their timestamp and use the first five days as the training set, the sixth day as the validation set, and the last day as the test set.
    \item Ouyang et al. \cite{ouyang2022learning} (DUGN) filters out users with less than 50 usage records, and apps with less than 5 usage records. If the time interval between two consecutive app usage records is less than 300 seconds, they are grouped into a session. App usage sessions from the first 6 consecutive days are used for training, and sessions from the last day are used for testing.
\end{itemize}

To conduct a rigorous and effective comparison with the methods presented in these two papers, we have separately adopted these two processing approaches for comparative experiments. Moreover, given that the data cleaning and partitioning method from PAULCI\cite{solomon2022predicting} has been more widely adopted in app prediction research, we decided to use this method for subsequent ablation studies and comparative analyses.

\section{Result}
In this section, we first conducted experiments comparing with baselines. Then, we performed ablation experiments on the different components of the model. Next, we compared the time encoding and fusion methods with traditional approaches. Subsequently, we replaced the components within the feature fusion module in Appformer and conducted comparative experiments. Following that, we replaced the Feature Fusion Module in Appformer and conducted a comparison experiment on the feature extraction module. Finally, we conducted clustering experiments on the POI data, including experiments to determine the number of clustering centers and selecting the best clustering method based on the number of centers.

\begin{table}[!t]
\centering
\caption{Performance comparison of our method with other methods based on PAULCI\cite{solomon2022predicting}.}
\label{tab_3}
\begin{threeparttable}
\begin{tabular}{lccccc}
\toprule
Methods & Hit@1 & Hit@3 & Hit@5 & MRR\tnote{1} & F1 \\
\midrule
Random forest\cite{breiman2001random} & 0.0004 & 0.0329 & 0.0642 & 0.0345 & 0.0012 \\
Logistic regression\cite{werbos1974beyond} & 0.0006 & 0.0763 & 0.0962 & 0.0528 & 0.0010 \\
XGBoost\cite{chen2016xgboost} & 0.0022 & 0.0206 & 0.0550 & 0.0068 & 0.0013 \\
MLP & 0.1156 & 0.2069 & 0.2704 & 0.1962 & 0.0085 \\
LSTM\cite{hochreiter1997long} & 0.1054 & 0.1763 & 0.2265 & 0.1735 & 0.0011 \\
GRU\cite{Cho2014LearningPR} & 0.1049 & 0.1845 & 0.2396 & 0.1773 & 0.0011 \\
PCA and LSTM\cite{xu2018predicting} & 0.1214 & 0.2069 & 0.2599 & 0.1955 & 0.0019 \\
NAP\cite{moreira2020nap} & 0.1058 & 0.1879 & 0.2416 & 0.1814 & 0.0012 \\
AppUsage2Vec\cite{zhao2019appusage2vec} & 0.1830 & 0.3470 & 0.4306 & 0.3026 & 0.0040 \\
DeepApp\cite{xia2020deepapp} & 0.2075 & 0.3854 & 0.4819 & 0.3351 & 0.0416 \\
PAULCI\cite{solomon2022predicting} & 0.2717 & 0.4753 & 0.5732 & 0.4098 & 0.1417 \\
Appformer + K\tnote{2} + T\tnote{3} & \textbf{0.3192} & \textbf{0.5213} & \textbf{0.6105} & \textbf{0.4511} & \textbf{0.1633} \\
\bottomrule
\end{tabular}
\begin{tablenotes}
    \item[1] The MRR represents the average reciprocal rank of the first correct answer in all results.
    \item[2] The K is K-Modes.
    \item[3] The T is Time encoding.
\end{tablenotes}
\end{threeparttable}
\end{table}

\begin{table*}[!t]
\centering
\caption{Performance comparison of our method with other methods based on DUGN\cite{ouyang2022learning}.}
\label{tab_x}
\begin{threeparttable}
\resizebox{\textwidth}{!}{%
\begin{tabular}{lcccccccccccccccc}
\toprule
Methods & \multicolumn{5}{c}{Hit@k} & \multicolumn{5}{c}{NDCG@k} & \multicolumn{5}{c}{MRR@k} \\
\cmidrule(r){2-6} \cmidrule(r){7-11} \cmidrule(r){12-16}
& 1 & 2 & 3 & 4 & 5 & 1 & 2 & 3 & 4 & 5 & 1 & 2 & 3 & 4 & 5 \\
\midrule
MRU\cite{shin2012understanding} & 0.2398 & 0.4196 & 0.5135 & 0.5670 & 0.5938 & 0.2398 & 0.3501 & 0.3921 & 0.4163 & 0.4299 & 0.2398 & 0.3210 & 0.3522 & 0.3668 & 0.3754 \\
MFU\cite{shin2012understanding} & 0.2472 & 0.4205 & 0.5073 & 0.5621 & 0.5987 & 0.2472 & 0.3565 & 0.3999 & 0.4235 & 0.4377 & 0.2472 & 0.3338 & 0.3628 & 0.3765 & 0.3838 \\
BPRMF\cite{10.5555/1795114.1795167} & 0.3293 & 0.4437 & 0.5188 & 0.5681 & 0.6077 & 0.3293 & 0.4015 & 0.4390 & 0.4602 & 0.4755 & 0.3293 & 0.3865 & 0.4115 & 0.4238 & 0.4317 \\
GRU4Rec\cite{hidasi2015session} & 0.3137 & 0.4493 & 0.5425 & 0.6028 & 0.6494 & 0.3137 & 0.3993 & 0.4459 & 0.4718 & 0.4899 & 0.3137 & 0.3815 & 0.4126 & 0.4277 & 0.4370 \\
AppUsage2Vec\cite{zhao2019appusage2vec} & 0.3333 & 0.4592 & 0.5436 & 0.6080 & 0.6560 & 0.3333 & 0.4127 & 0.4549 & 0.4827 & 0.5012 & 0.3333 & 0.3962 & 0.4244 & 0.4405 & 0.4501 \\
SR-GNN\cite{wu2019session} & 0.3342 & 0.4716 & 0.5563 & 0.6154 & 0.6626 & 0.3342 & 0.4209 & 0.4632 & 0.4887 & 0.5070 & 0.3342 & 0.4029 & 0.4311 & 0.4459 & 0.4554 \\
DUGN\cite{ouyang2022learning} & 0.3479 & 0.4768 & 0.5593 & 0.6215 & 0.6710 & 0.3479 & 0.4292 & 0.4705 & 0.4973 & 0.5164 & 0.3479 & 0.4124 & 0.4399 & 0.4554 & 0.4653 \\
Appformer + K + T & \textbf{0.4268} & \textbf{0.5550} & \textbf{0.6230} & \textbf{0.6656} & \textbf{0.6960} & \textbf{0.4268} & \textbf{0.5550} & \textbf{0.5979} & \textbf{0.6192} & \textbf{0.6323} & \textbf{0.4268} & \textbf{0.4909} & \textbf{0.5136} & \textbf{0.5242} & \textbf{0.5303} \\
\bottomrule
\end{tabular}%
}
\end{threeparttable}
\end{table*}

\subsection{Comparison with baselines}

The comparison results of our method with PAULCI\cite{solomon2022predicting} are shown in Table \ref{tab_3}, with the experimental results of other methods in the table also sourced from PAULCI. The comparison results with DUGN\cite{ouyang2022learning} are shown in Table \ref{tab_x}, and the experimental results of other methods in that table are also derived from DUGN. Across all metrics, our method achieves SOTA performance, illustrating its effectiveness in the field of mobile app prediction.

The significance of user information, spatio-temporal information, and historical app sequences in mobile app prediction is self-evident. Previous research has gradually incorporated these types of data, continuously enhancing their representations. Coupled with advancements in feature extraction techniques, this has led to a steady improvement in performance metrics.

For instance, the PAULCI method employs graph embeddings to depict spatial context and multi-modal embeddings to represent temporal context, user identifiers, and previously utilized applications. It also uses a deep learning framework, which includes GRU, attention layers, and softmax layers, for feature extraction and prediction, yielding good results. The DUGN method utilizes a dynamic graph structure, allowing it to capture the intricate correlations between various applications and track the evolution of user interests. By integrating multiple interests, it generates comprehensive user embeddings that serve as a robust foundation for future app recommendations, leading to commendable outcomes.

In our research, we implemented K-Modes clustering on POI data and improved the time encoding and fusion methods to enhance the expressiveness of the fundamental data. Furthermore, our proposed Appformer uniquely merges different types of data and enhances feature extraction, thereby leading to superior results compared to the other methods.

\begin{table}[!t]
\centering
\caption{Results for module ablation.}
\label{tab_4}
\begin{tabular}{lccccc}
\toprule
Methods & Hit@1 & Hit@3 & Hit@5 & MRR & F1 \\
\midrule
Appformer & 0.3110 & 0.5137 & 0.6045 & 0.4437 & 0.1543 \\
Appformer + T & 0.3173 & 0.5196 & 0.6064 & 0.4486 & 0.1608 \\
Appformer + K & 0.3171 & 0.5201 & 0.6101 & 0.4497 & 0.1620 \\
Appformer + K + T & \textbf{0.3192} & \textbf{0.5213} & \textbf{0.6105} & \textbf{0.4511} & \textbf{0.1633} \\
\bottomrule
\end{tabular}
\end{table}

\subsection{Module ablation study}
We performed a comprehensive ablation study of our approach to scrutinize its key components. Initially, the complete model (Appformer+K+T) was employed, yielding optimal performance, refer to Table \ref{tab_4}. Subsequently, we systematically removed its components and conducted experiments under three different configurations.

In the first experiment (Appformer+K), we excised the handling and integration (T) of time information, which led to a performance degradation. This suggests that our handling and integration operations for time information positively contribute to the model's overall effectiveness. In the second experiment (Appformer+T), we omitted the clustering operation (K) of the POI data, also leading to a performance degradation. This implies that the clustering operation of  POI data plays a significant role in enhancing spatial information.

Lastly, in the third experiment (Appformer), the Appformer model still surpassed the baseline performance, even after removing these two critical components. This reaffirms the uniqueness and efficacy of the Appformer framework.

\begin{table}[!t]
\centering
\caption{Results for time encoding method comparison experiment.}
\label{tab_5}
\begin{tabular}{lccccc}
\toprule
Methods & Hit@1 & Hit@3 & Hit@5 & MRR & F1 \\
\midrule
Appformer + $\mathrm{T}_{\mathrm{1}}$\cite{zhou2021informer} & 0.3057 & 0.5056 & 0.5927 & 0.4365 & 0.1492 \\
Appformer + T & \textbf{0.3173} & \textbf{0.5196} & \textbf{0.6064} & \textbf{0.4486} & \textbf{0.1608} \\
\bottomrule
\end{tabular}
\end{table}

\begin{table}[!t]
\centering
\caption{Results for ablation experiment of Appformer Decoder input.}
\label{tab_6}
\begin{threeparttable}
\begin{tabular}{lccccc}
\toprule
Methods & Hit@1 & Hit@3 & Hit@5 & MRR & F1 \\
\midrule
AKT\tnote{1} (Decoder(l+t+u)) & 0.3180 & 0.5207 & 0.6083 & 0.4497 & 0.1598 \\
AKT(Decoder(u+t)) & 0.3161 & 0.5192 & 0.6075 & 0.4481 & 0.1605 \\
AKT(Decoder(u+l)) & 0.3170 & 0.5198 & 0.6079 & 0.4490 & 0.1590 \\
AKT(Decoder(t+l)) & 0.3186 & 0.5209 & 0.6091 & 0.4504 & 0.1629 \\
AKT(Decoder(u)) & 0.3159 & 0.5169 & 0.6046 & 0.4471 & 0.1578 \\
AKT(Decoder(t)) & 0.3176 & 0.5207 & 0.6089 & 0.4496 & 0.1619 \\
AKT(Decoder(l)) & \textbf{0.3192} & \textbf{0.5213} & \textbf{0.6105} & \textbf{0.4511} & \textbf{0.1633} \\
\bottomrule
\end{tabular}
\begin{tablenotes}
    \item[1] The AKT is Appformer + K +T.
\end{tablenotes}
\end{threeparttable}
\end{table}

\subsection{Time encoding comparison experiment}
\label{Time encoding comparison experiment}
Under the premise that all basic prediction models adopt Appformer, we conducted a comparative experiment to contrast the improved time encoding and fusion methods with the basic time encoding and addition method $\mathrm{T}_{\mathrm{1}}$\cite{zhou2021informer}. As shown in Table \ref{tab_5}, the experimental results clearly demonstrate that our proposed improved method outperforms the basic method across various metrics. This indicates that our method accurately captures the inherent patterns of time information, enabling the model to better understand the changes in users' app usage habits over time, thereby enhancing prediction accuracy and stability.

\subsection{Ablation experiment of Appformer Decoder input}
\label{Ablation experiment of Appformer decoder input}

Our Appformer uses the classic Encoder-Decoder architecture in the feature extraction module, which is a common approach in time series prediction. The role of the Encoder is to transform the input sequence into a high-dimensional representation, while the Decoder decodes this high-dimensional representation and combines it with an additional conditional vector to generate the target sequence. Therefore, we feed all data into the Encoder, and the decoder only needs the appropriate data.

We conducted ablation experiments specifically on the input to the Decoder, with the results presented in Table \ref{tab_6}. The optimal results were obtained when only POI data was input into the Decoder. This is because the user sequence itself consists of repetitive numbers, and the time series changes slowly. The presence of redundant data could potentially interfere with the Decoder's decoding operation. Hence, we consider AKT (Decoder(l)) as our optimal result. The Appformer + K + T mentioned in previous experiments refers to this AKT (Decoder(l)).

\begin{table}[!t]
\centering
\caption{Results for Feature Extraction Module comparison experiment.}
\label{tab_7}
\begin{tabular}{lccccc}
\toprule
Methods & Hit@1 & Hit@3 & Hit@5 & MRR & F1 \\
\midrule
AKT(Auto\cite{wu2021autoformer}) & 0.3084 & 0.5062 & 0.5899 & 0.4369 & 0.1574 \\
AKT(FED\cite{zhou2022fedformer}) & 0.3096 & 0.5070 & 0.5900 & 0.4378 & 0.1588 \\
AKT(Quat\cite{chen2022learning}) & 0.1883 & 0.3327 & 0.4089 & 0.2947 & 0.0413 \\
AKT & \textbf{0.3192} & \textbf{0.5213} & \textbf{0.6105} & \textbf{0.4511} & \textbf{0.1633} \\
\bottomrule
\end{tabular}
\end{table}

\subsection{Feature Extraction Module comparison experiment}
\label{Feature extraction module comparison experiment}
We selected equivalent modules from AutoFormer \cite{wu2021autoformer}, FEDformer \cite{zhou2022fedformer}, and Quatformer \cite{chen2022learning} to replace the components in the Feature Extraction Module of our Appformer, and conducted a comprehensive comparative experiment. Specifically, for AutoFormer, we introduced the series decomposition block, aiming to decompose the time series into trend-cyclical and seasonal parts. Simultaneously, we replaced the Cross-Modal Data Fusion Module with the auto-correlation mechanism. For FEDformer, we incorporated the mixture of expert decomposition block (MOEDecomp), which utilizes average filters of different sizes and data-related weights to extract multiple trend components from the input signal. Additionally, we utilized fourier enhanced block instead of the Cross-Modal Data Fusion Module. Regarding Quatformer, we employed decoupling learning-to-rotate attention to replace the Cross-Modal Data Fusion Module. This module utilizes quaternions to handle periodicity information and employs a global memory decoupling approach to handle complex periodic patterns in time series. Furthermore, we used trend normalization instead of layer normalization to normalize sequential representations in the model's hidden layers.

As shown in the experimental results in Table \ref{tab_7}, we carefully selected and made reasonable replacements and adjustments to these modules. However, they were primarily designed for long sequence prediction tasks. Therefore, they might not have fully adapted to the characteristics of our short sequence prediction task. This could explain why the results did not meet our expectations.

\begin{figure*}[!t]
\centering
\begin{minipage}{0.5\textwidth}
  \centering
  \includegraphics[width=\linewidth]{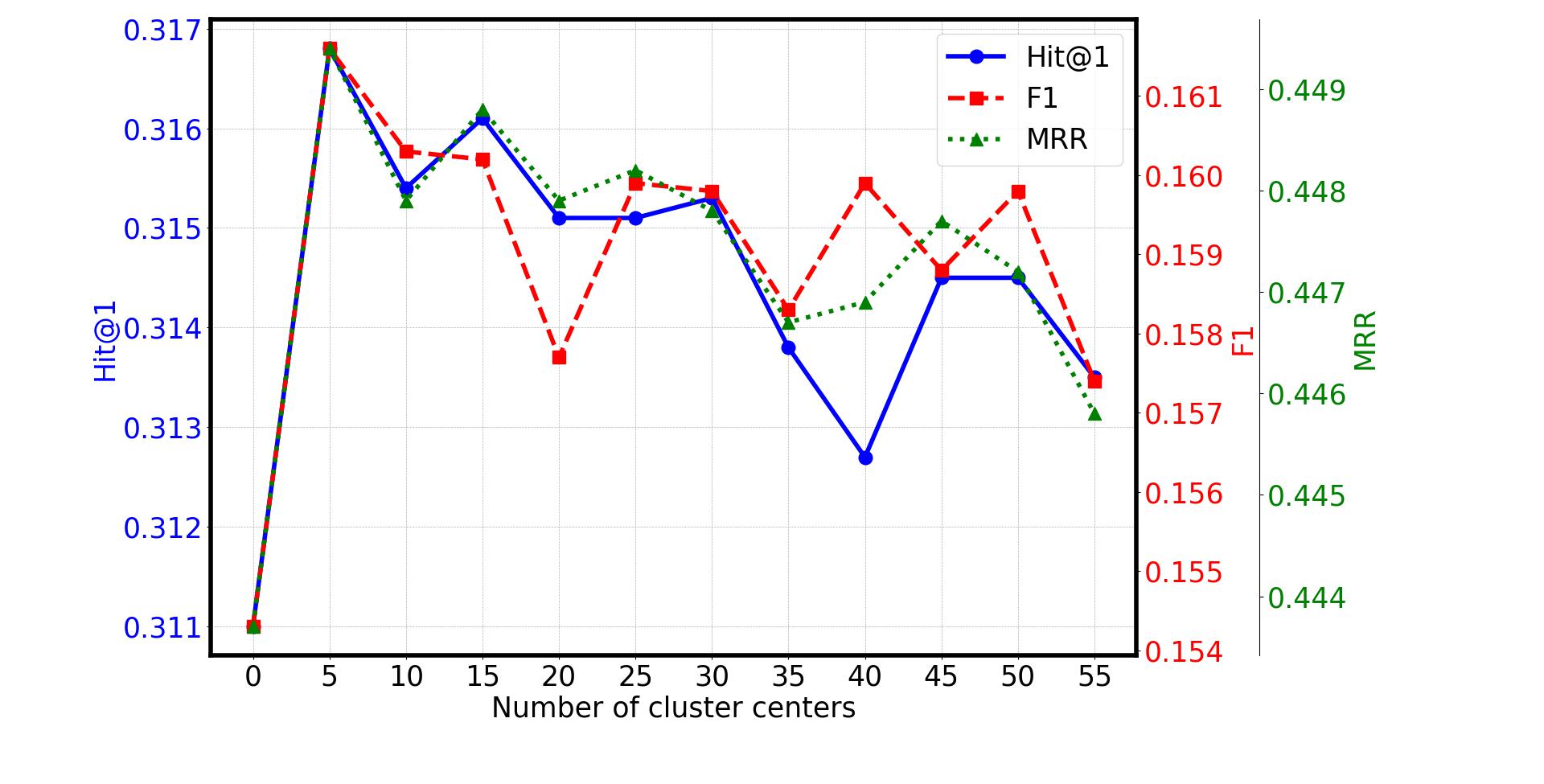}
  \caption{Performance trends with different cluster sizes.}
  \label{fig_5}
\end{minipage}\hfill
\begin{minipage}{0.5\textwidth}
  \centering
  \includegraphics[width=\linewidth]{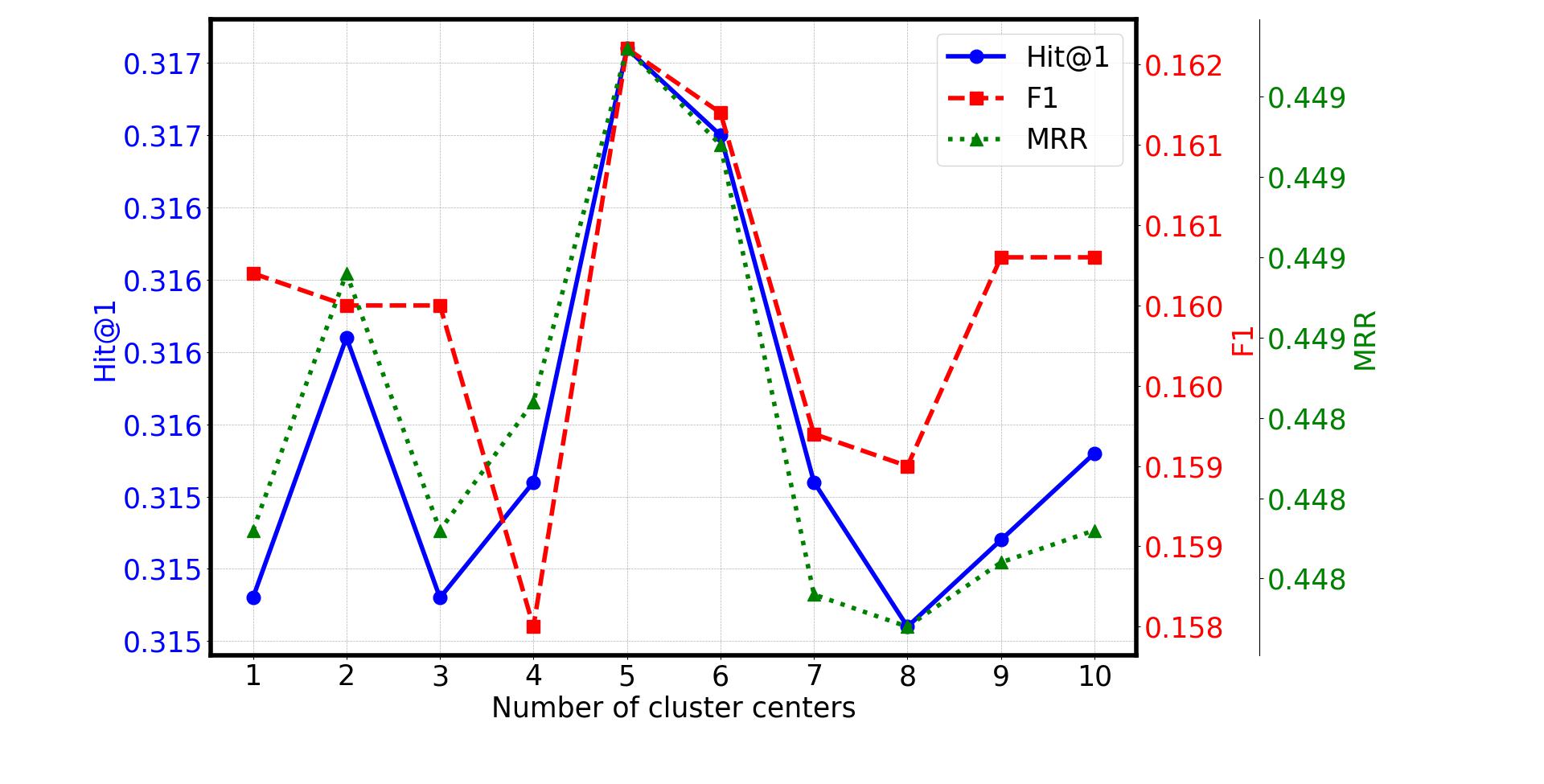}
  \caption{Detailed trends for cluster sizes 0-10.}
  \label{fig_6}
\end{minipage}
\end{figure*}

\begin{table}[!t]
\centering
\caption{Results of clustering experiments with different cluster sizes.}
\label{tab_8}
\begin{tabular}{cccccc}
\toprule
Number of centers & Hit@1 & Hit@3 & Hit@5 & MRR & F1 \\
\midrule
0 & 0.3110 & 0.5137 & 0.6045 & 0.4437 & 0.1543 \\
5 & \textbf{0.3168} & \textbf{0.5201} & \textbf{0.6097} & \textbf{0.4494} & \textbf{0.1616} \\
10 & 0.3154 & 0.5190 & 0.6078 & 0.4479 & 0.1603 \\
15 & 0.3161 & 0.5197 & 0.6092 & 0.4488 & 0.1602 \\
20 & 0.3151 & 0.5193 & 0.6076 & 0.4479 & 0.1577 \\
25 & 0.3151 & 0.5200 & 0.6078 & 0.4482 & 0.1599 \\
30 & 0.3153 & 0.5184 & 0.6075 & 0.4478 & 0.1598 \\
35 & 0.3138 & 0.5175 & 0.6074 & 0.4467 & 0.1583 \\
40 & 0.3127 & 0.5192 & 0.6089 & 0.4469 & 0.1599 \\
45 & 0.3145 & 0.5193 & 0.6075 & 0.4477 & 0.1588 \\
50 & 0.3145 & 0.5182 & 0.6062 & 0.4472 & 0.1598 \\
55 & 0.3135 & 0.5155 & 0.6049 & 0.4458 & 0.1574 \\
\bottomrule
\end{tabular}
\end{table}

\begin{table}[!t]
\centering
\caption{Experimental results for cluster size 0-10.}
\label{tab_9}
\begin{tabular}{cccccc}
\toprule
Number of centers & Hit@1 & Hit@3 & Hit@5 & MRR & F1 \\
\midrule
1 & 0.3149 & 0.5190 & 0.6081 & 0.4479 & 0.1602 \\
2 & 0.3158 & 0.5204 & 0.6096 & 0.4487 & 0.1600 \\
3 & 0.3149 & 0.5190 & 0.6086 & 0.4479 & 0.1600 \\
4 & 0.3153 & 0.5192 & 0.6088 & 0.4483 & 0.1580 \\
5 & \textbf{0.3168} & \textbf{0.5201} & \textbf{0.6097} & \textbf{0.4494} & \textbf{0.1616} \\
6 & 0.3165 & 0.5200 & 0.6089 & 0.4491 & 0.1612 \\
7 & 0.3153 & 0.5183 & 0.6077 & 0.4477 & 0.1592 \\
8 & 0.3148 & 0.5183 & 0.6084 & 0.4476 & 0.1590 \\
9 & 0.3151 & 0.5189 & 0.6080 & 0.4478 & 0.1603 \\
10 & 0.3154 & 0.5190 & 0.6078 & 0.4479 & 0.1603 \\
\bottomrule
\end{tabular}
\end{table}

\begin{figure*}[!t]
\centering
\begin{minipage}{0.5\textwidth}
  \centering
  \includegraphics[width=\linewidth]{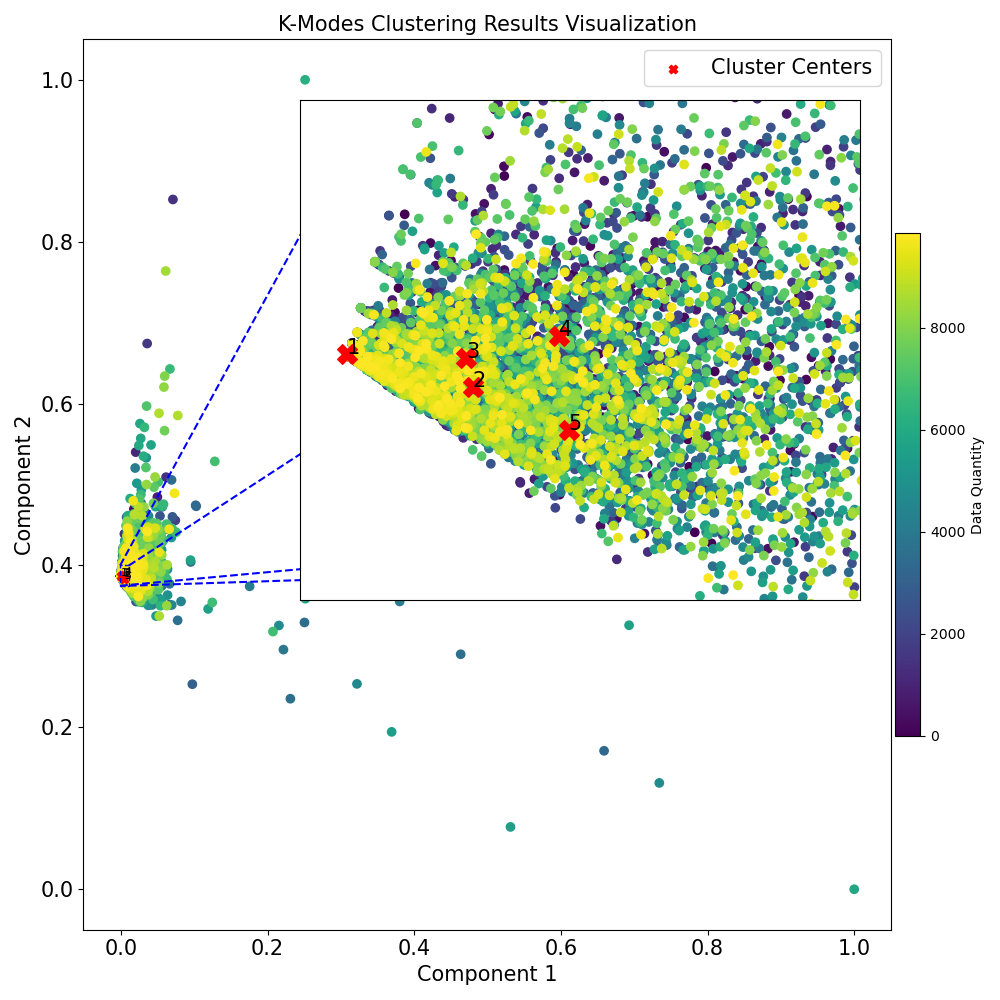}
  \caption{K-Modes Visualization.}
  \label{fig_7}
\end{minipage}\hfill
\begin{minipage}{0.5\textwidth}
  \centering
  \includegraphics[width=\linewidth]{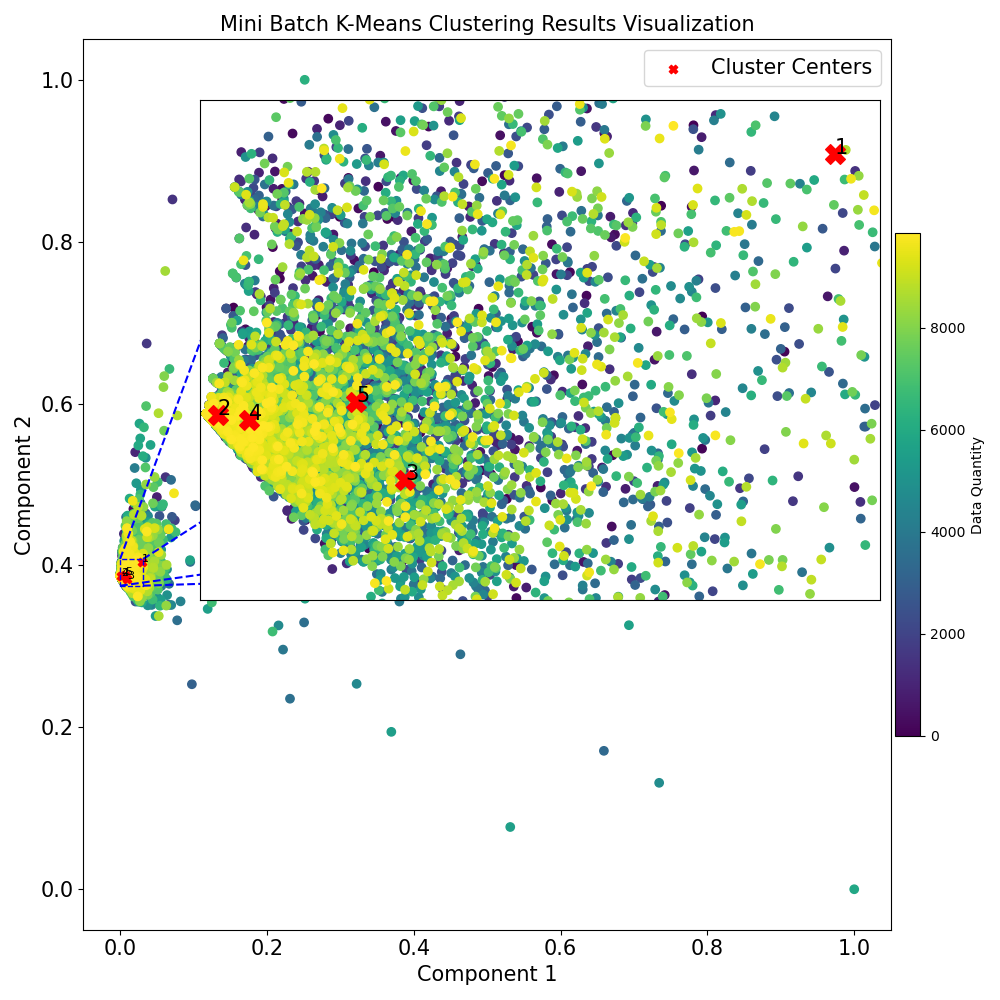}
  \caption{Mini Batch K-Means Visualization.}
  \label{fig_8}
\end{minipage}\hfill
\\
\begin{minipage}{0.333333333\textwidth}
  \centering
  \includegraphics[width=\linewidth]{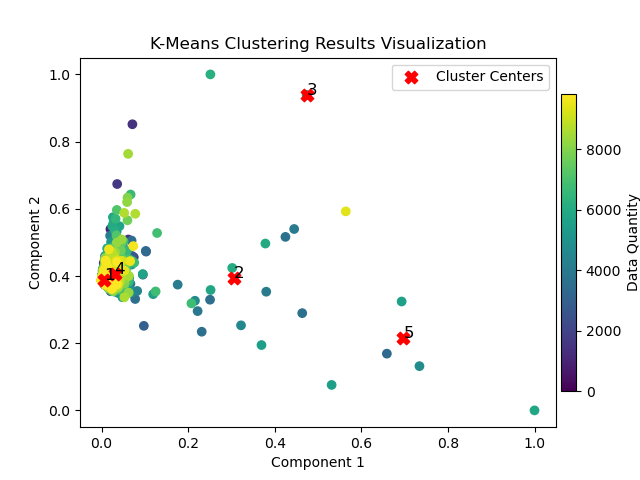}
  \caption{K-Means Visualization.}
  \label{fig_9}
\end{minipage}\hfill
\begin{minipage}{0.333333333\textwidth}
  \centering
  \includegraphics[width=\linewidth]{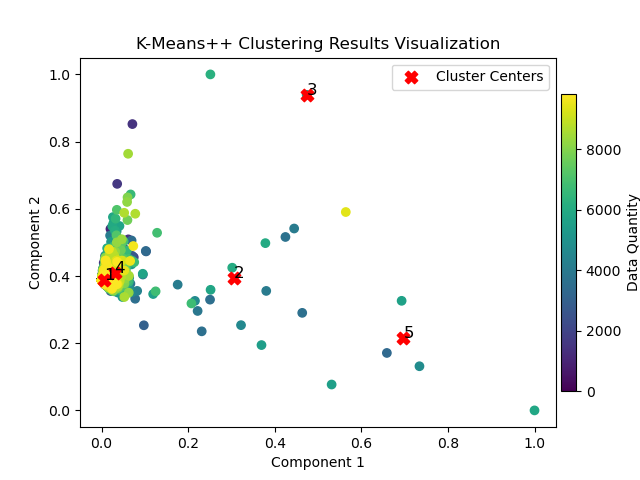}
  \caption{K-Means++ Visualization.}
  \label{fig_10}
\end{minipage}\hfill
\begin{minipage}{0.333333333\textwidth}
  \centering
  \includegraphics[width=\linewidth]{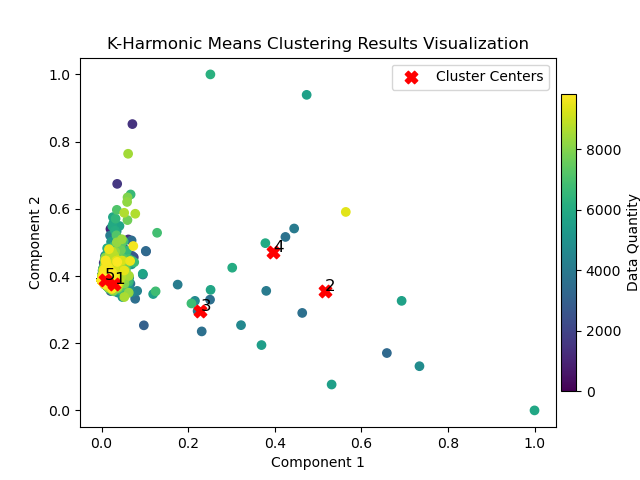}
  \caption{K-Harmonic Means Visualization.}
  \label{fig_11}
\end{minipage}
\end{figure*}

\subsection{Clustering experiment on the POI data}
\label{Clustering experiments}

As outlined in Section \ref{Data preprocessing}, our experimental design includes a comparison and analysis of various clustering algorithms, such as K-Means, Mini Batch K-Means, K-Means++, K-Modes, and K-Harmonic Means. The initial step involves determining the number of cluster centers, followed by evaluating the clustering methods with this established number. The necessity of this process stems from the effectiveness of clustering POI data, which guides our selection of cluster quantities and methods. Despite minor differences in performance indicators among the methods, this phase is crucial. It allows us to identify the most suitable clustering approach for our data, reinforcing the value of our methodical exploration in understanding POI data's underlying patterns.

\textbf{Determining the number of cluster centers:} Since other clustering algorithms are evolved or improved based on K-Means, we used the most basic K-Means algorithm in combination with Appformer for the cluster center comparison experiment. 

The experimental results in Tables \ref{tab_8} and \ref{tab_9} are supplemented by the visualizations in Figures \ref{fig_5} and \ref{fig_6}. The outcomes in Tables \ref{tab_8} and \ref{tab_9} demonstrate that the optimal number of clusters for POI data, based on Hit@1, Hit@3, Hit@5, MRR, and F1 score, is 5. However, the relationship between the number of clusters and performance is not simply linear. As shown in Figure \ref{fig_5}, performance generally declines as the number of clusters increases from 5 to 55, albeit with some fluctuations. Figure \ref{fig_6} provides a more detailed view for cluster sizes 0-10, confirming that performance peaks at 5 clusters.

Although the correlation between the number of clusters and performance is not very strong, these experiments empirically demonstrate that 5 serves as a balancing point, where there are enough clusters to meaningfully group similar items while avoiding excessive fragmentation. The initial performance improvement from 0 to 5 clusters illustrates the benefits of clustering, but the decline beyond 5 clusters suggests that too many clusters could be detrimental, possibly because it makes each cluster too small and too specific.

\textbf{Testing different clustering methods:} Experimental results, as shown in Table \ref{tab_10}, indicate that when considering three key metrics: Hit@1, MRR, and F1 score, the K-Modes algorithm demonstrates superior performance. To further understand these results, we employed Principal Component Analysis (PCA) to reduce the dimensionality of the POI vectors corresponding to base station IDs and the cluster center vectors obtained by different clustering algorithms down to two dimensions for a detailed visual analysis. The visualization figures are presented from Figures \ref{fig_7}, \ref{fig_8}, \ref{fig_9}, \ref{fig_10}, and \ref{fig_11}.

The visual analysis revealed that some data points significantly deviate from the central region of the overall data, identified as noise points. Notably, algorithms such as K-Means, K-Means++, and K-Harmonic Means incorrectly identify these noise points as cluster centers, which is suboptimal. In contrast, the Kmodes and Mini Batch K-Means algorithms can effectively recognize and disregard these noise points, assigning them to the nearest cluster, thereby minimizing the adverse impact of noise on the clustering outcome.

Further examination of the distribution of cluster centers for these two algorithms shows that some cluster centers of Mini Batch K-Means are located in sparser areas of the data, while the K-Modes algorithm’s cluster centers are evenly distributed in denser data regions. Considering the magnitude difference between dense and sparse areas, this indicates that the K-Modes algorithm is more rational and effective in selecting cluster centers, ensuring that they are located in the true core areas of the data, thus enhancing the accuracy and reliability of the clustering results.

In summary, through the clustering analysis of POI vectors, the K-Modes algorithm exhibits significant advantages in handling noise, maintaining data centrality, and improving clustering accuracy. These findings not only validate the effectiveness of our experimental results but also provide valuable references for future clustering analyses on similar datasets.

\begin{table}[!t]
\centering
\caption{Results for different clustering methods.}
\label{tab_10}
\begin{threeparttable}
\resizebox{0.5\textwidth}{!}{%
\begin{tabular}{lccccc}
\toprule
Methods & Hit@1 & Hit@3 & Hit@5 & MRR & F1 \\
\midrule
A\tnote{1} + K-Means & 0.3168 & 0.5201 & 0.6097 & 0.4494 & 0.1616 \\
A + Mini Batch K-Means & 0.3166 & 0.5194 & 0.6089 & 0.4489 & 0.1615 \\
A + K-Means++ & 0.3164 & 0.5199 & 0.6097 & 0.4492 & 0.1613 \\
A + K-Harmonic Means & 0.3164 & 0.5195 & 0.6091 & 0.4491 & 0.1598 \\
A + K-Modes & \textbf{0.3171} & \textbf{0.5201} & \textbf{0.6101} & \textbf{0.4497} & \textbf{0.1620} \\
\bottomrule
\end{tabular}%
}
\begin{tablenotes}
    \item[1] The A is Appformer.
\end{tablenotes}
\end{threeparttable}
\end{table}

\section{Conclusion}

This study showcases Appformer as a novel approach for predicting mobile app usage, demonstrating its potential through innovative multi-modal data fusion, feature extraction strategies, and the use of Transformer-like architecture. Specifically, we enhanced the preprocessing of location and time data, laying a solid foundation for data fusion and feature extraction. Through rigorous experimental validation, Appformer outperforms existing methods on several predictive performance metrics, such as Hit@1, MRR, and F1 score. This highlights Appformer's robustness and efficiency in predicting mobile app usage, showcasing its potential to improve personalized services and enhance user experience.

In our in-depth analysis of the Appformer approach, we identified its limitations as involving challenges in the data fusion process, constraints in feature extraction, and difficulties in model updates and maintenance. Specifically, effectively integrating multimodal data, addressing inconsistencies and conflicts within the data, and avoiding the introduction of bias or loss of important information during data fusion are key challenges. Additionally, the model may overly rely on predefined features and extraction methods, limiting its ability to explore intrinsic patterns in the data, especially when dealing with complex or unstructured data. Moreover, to adapt to the constantly evolving patterns of app usage and the emergence of new applications, the model requires regular updates. However, frequent updates may demand substantial computational resources and cause short-term fluctuations in model performance. Maintaining the model's continuous learning and adaptation to new data patterns while avoiding the disruption of existing knowledge poses a significant challenge. These multifaceted challenges not only deepen our understanding of the limitations of the Appformer method but also point out directions for future research, namely improving the applicability, performance, and adaptability of the model in the dynamic field of app usage prediction. This ensures that it can effectively adapt to the ever-changing digital ecosystem and meet the diverse needs of users.

We would like to release our source code.

\bibliographystyle{IEEEtran}
{\footnotesize
\bibliography{appformer}
}

\end{document}